\documentclass[letterpaper, 10 pt, conference]{ieeeconf}

\IEEEoverridecommandlockouts
\usepackage{algorithm}      
\usepackage{algpseudocode}  
\usepackage{amsmath}
\usepackage{amsfonts}    
\usepackage{graphicx}
\usepackage{multicol}
\usepackage{multirow}
\usepackage{subcaption}
\usepackage{todonotes}
\usepackage{units}
\usepackage[bookmarks=true]{hyperref}

\newcommand{\R}{\mathbb{R}}
\algnewcommand\algorithmicswitch{\textbf{switch}}
\algnewcommand\algorithmiccase{\textbf{case}}
\algnewcommand\algorithmicdefault{\textbf{default}}
\algdef{SE}[SWITCH]{Switch}{EndSwitch}[1]{\algorithmicswitch\ #1\ \algorithmicdo}{\algorithmicend\ \algorithmicswitch}%
\algdef{SE}[CASE]{Case}{EndCase}[1]{\algorithmiccase\ #1}{\algorithmicend\ \algorithmiccase}%
\algdef{SE}[DEFAULT]{Default}{EndDefault}[1]{\algorithmicdefault\ #1}{\algorithmicend\ \algorithmicdefault}
\algtext*{EndCase}%
\algtext*{EndDefault}%

\graphicspath{{img/}}

\title{\LARGE \bf SwarmMesh:\\
  A Distributed Data Structure\\
  for Cooperative Multi-Robot Applications}

\author{%
  \authorblockN{%
    Nathalie~Majcherczyk~\IEEEmembership{Student~Member,~IEEE}
    and
    Carlo Pinciroli~\IEEEmembership{Member,~IEEE}}
  \authorblockA{Robotics Engineering, Worcester Polytechnic Institute, MA, USA\\
  Email: \{nmajcher,cpinciroli\}@wpi.edu}
}

\begin{document}

\maketitle

\begin{abstract}
  We present an approach to the distributed storage of data across a swarm of mobile robots that forms a shared global memory. We assume that external storage infrastructure is absent, and that each robot is capable of devoting a quota of memory and bandwidth to distributed storage. Our approach is motivated by the insight that in many applications data is collected at the periphery of a swarm topology, but the periphery also happens to be the most dangerous location for storing data, especially in exploration missions. Our approach is designed to promote data storage in the locations in the swarm that best suit a specific feature of interest in the data, while accounting for the constantly changing topology due to individual motion. We analyze two possible features of interest: the data type and the data item position in the environment. We assess the performance of our approach in a large set of simulated experiments. The evaluation shows that our approach is capable of storing quantities of data that exceed the memory of individual robots, while maintaining near-perfect data retention in high-load conditions.
\end{abstract}

\section{Introduction}
\label{sec:introduction}
Recent work studies with the integration of multi-robot systems with centralized
computation platforms, such as databases or cloud computing systems. This
approach enables one to aggregate information in a central location and perform
efficient map merging, task allocation, and global state estimation --- in other
words, combining data storage with computational capabilities. This approach is
particularly effective in indoor environments, such as warehouses, production
chains, and hospitals, in which communication with a central system can be
expected to be reliable.

However, many applications are not easily amenable to this approach. Mapping in
remote locations, space applications, and disaster recovery are examples in
which access to a centralized infrastructure is problematic, heavily limited, or
even impossible. In these applications, rather than envisioning a multi-robot
system as \emph{part} of a larger infrastructure, it would be desirable for it
to \emph{be} the infrastructure. These applications also entail the collection
of large amounts of data, whose storage might exceed the capacity of any
individual robot.

As a step in this direction, we study the realization of a decentralized data
structure for storing, managing, and performing computation with shared data. We
make three basic assumptions:
\begin{itemize}
\item Every robot devotes a quota of memory and bandwidth to storing and routing
  data. The amount of memory can change across robots;
\item The amount of data that the robots must store is larger than the memory
  capacity of any individual robot;
\item The network topology is dynamic due to robot motion.
\end{itemize}

Given these assumptions, we study how to distribute the data across the
swarm. In designing a solution, we realized that in many applications certain
features of the data play an important role for mission success. For example,
mission-critical data should be stored in well-connected robots---in case of a
temporary disconnection this data would be as widely available as
possible. Analogously, the physical location of the data might suggest that
certain robots are more suitable for storage than others.

The rest of this paper is organized as follows. In Section~\ref{sec:relatedwork}
we discuss related work. The design of our data structure is presented in
Section~\ref{sec:methodology}. We report the results of our performance
evaluation in Section~\ref{sec:evaluation}, and conclude the paper in
Section~\ref{sec:conclusions}.

\section{Related Work}
\label{sec:relatedwork}

In peer-to-peer networks, common implementations of data sharing involve
Distributed Hash Tables (DHTs). DHTs couple a distributed key partitioning
algorithm and a structured overlay network to provide a self-organized data
storage service. Information is abstracted in the form of tuples which are
\texttt{(key, value)} pairs. The fundamental problem is to decide how to
distribute tuples between nodes for storage. The key partitioning algorithm
assigns ownership of a set of keys to each node in the network. The overlay
network imposes a routing structure that makes for efficient search across the
nodes. Comparative surveys~\cite{Urdaneta2011},\cite{Lua2004} highlight the main
features of these protocols. The Content-Addressable Network (CAN) protocol
partitions the key space by splitting a virtual toroidal space into zones. CAN
maps tuples to points owned by nodes in the virtual space using a uniform hash
function~\cite{ratnasamy2001scalable}. In the Chord~\cite{stoica2001chord},
Pastry~\cite{rowstron2001pastry}, and Tapestry~\cite{zhao2004tapestry}
protocols, nodes determine NodeIDs according to the structure of the desired
overlay network. Tuples are then addressed directly to NodeIDs or partial
NodeIDs. These distributed data structures provide self-organizing, scalable and
addressable storage. However, node additions and removals are costly as the
topology needs to be maintained through reorganization. Furthermore, they can
cause local network failures.  Because they form relations between nodes
randomly, unstructured overlay networks such as Gnutella and
BitTorrent~\cite{cohen2003bittorrent} provide alternatives when the network
participant turnover is high. These protocols offer robustness to node removals
at the cost of increased degree of centralization or loss of guarantees when
locating data. 

In the above mentioned protocols, the selection of neighboring nodes in the
overlay network lacks physical meaning. This means that neighbors in the network
could be far away from each other. Since routing information over longer
geographical distances increases energy consumption and latency, there has been
an effort to incorporate node location into overlay networks. Three main trends
exist within this body of research: (1) Geographic layout, which constructs the
overlay network so that neighbors are close in the physical
space~\cite{Ahullo2008},~\cite{Wu},
~\cite{Pethalakshmi2014},~\cite{Matsuura2007}; (2) Proximity routing, which
considers node proximity while routing in the existing overlay network
\cite{Araujo}; (3) Proximity neighbor selection, which weighs in proximity
between neighbors when constructing the overlay
network~\cite{castro2002exploiting}. These methods add a notion of node
locality. However, the network topology only changes to accommodate node
additions and removals but not motion. Therefore, they fail to capture the
inherent dynamicity of robotic
systems. 



There is a vast body of research in Mobile Ad-Hoc Networks (MANETs) that seeks
to address communication between mobile interconnected devices. Some sensor
networks have motion and fall in that category. One trend in those systems has
been to use naming and data-centric routing and storage. This means that a name
is associated to given data and that name determines to which node the data is
addressed~\cite{Ratnasamy2003}. Similarly to swarm systems, the main features of
sensor networks are that they are limited in energy, memory and computational
power. Sensor networks perform distributed data processing and storage. However,
the goal is to eventually offload the processed data to a base
station. Furthermore, sensors typically do not act on the environment or perform
cooperative and autonomous decision-making.  Vehicular Ad Hoc Networks (VANETs)
are systems of interconnected cars and road stations. Different types of routing
protocols have been studied within that field, they can be divided into:
proactive routing, reactive routing and position-based routing which depends on
beaconing and forwarding~\cite{zeadally2012},
\cite{Yousefi2006},~\cite{Viswacheda2015}. These systems share some similarities
with swarms but differ in that they have specific topologies and mobility
patterns. Typically, cars in the back make decisions based on cars up front. The
lanes and roads are narrow so the number of direct neighbors is small.

Several papers compare and assess the use of existing databases in multi-robot
applications~\cite{Ravichandran2018},~\cite{fourie2017}
and~\cite{fiannaca2015}. These comparisons reveal that most existing databases
rely on a central server. An exception is the work of Sun \emph{et
  al.}~\cite{sun_decentralized_2010}, who adapted Distributed Heterogeneous Hash
Tables and position-based routing to propose a solution for task allocation in a
warehouse setting.
In the context of swarm robotics, Pinciroli \emph{et al.} proposed a distributed
tuple space called \textit{virtual stigmergy}~\cite{Pinciroli2018} that copes
with frequent topology changes. In this approach, each robot maintains a local
time-stamped copy of the data which is only accessed upon read and write
operations. This mechanism works well with node mobility and limited bandwidth
but it leads to full data duplication. This means that the collective memory of
the system is severaly under-utilized. The SOUL file sharing
protocol~\cite{varadharajan2019} builds on virtual stigmergy and unstructured
overlay networks to enable sharing of larger-size data in the form of
\texttt{(key, blob)} pairs. SOUL involves locally storing blob meta-data on each
node and splitting blobs into datagrams across different nodes. This
decomposition uses a bidding mechanisms that minimize the reconstruction cost at
so-called processor nodes. This method addresses the problem of managing data
files with a focus on how to split, distribute and recombine them. Memory usage
is improved but meta-data is still fully duplicated across nodes for each of the
files. Various update and bidding processes increase latency in the network.

In this paper, we take inspiration from existing methods and propose a novel
approach to distributed data sharing. Our design embraces the decentralized
nature of robot swarms and the constant change and volatility in the network
topology that results from robot motion. We organize our data flow based on
instantaneous local properties at each node so as to get a memory efficient,
consensus-free approach with a low communication overhead.

\section{Problem Statement}
\label{sec:probstat}
In this section, we describe the fundamental assumptions imposed both on the
multi-robot system and on the nature of the events to record in the physical
environment. We proceed by describing challenges of the distributed storage
problem in this context.

\subsection{Ad-hoc Robotic Network}

We consider an autonomous and decentralized system of $N$ robots which act as
both the infrastructure for storing information and sole users of this
information. We define the system across the following features:

\paragraph{Communication Modalities} We assume that the robots have the ability
to exchange data within a communication range $C$. This implies the existence of
an ad-hoc network with each robot acting as a node. We further assume that robot
communication is limited to gossiping, i.e., broadcasting messages to all
neighbors within $C$. Because we also desire to route some messages from one
robot to another using the point-to-point communication modality, we assume that
the robots have a constant unique identifier $i \in [1, N]$ and a variable node
identifier $\delta_i$ made known to their neighbors. The knowledge of $i$
singles out a specific robot, while $\delta_i$ enables the selection of a
suitable storage node for a specific tuple.
    
\paragraph{Finite Resources}
We impose a realistic finite bandwidth on outgoing messages. We also limit the
memory capacity $M_i$ of each robot allotted for the self-organizing data
management process. Variable $m_i(t)$ records the amount of memory used by robot
$i$ at a given time.

\paragraph{Dynamic Topology}
The robots are moving according to a logic defined by the developer. Robot
motion follows linear dynamics and has a limited speed. The number of neighbors
$ngbrs_i(t)$ of a robot changes over time.



\subsection{Inputs}
\label{ssec:events}
We consider inputs to the data structure to stem from events which have a
position $\mathbf{x} \in \R^d$ and happen at a time $t \in \mathbb{N}$. Such
events can be, e.g., records of a physical phenomenon sampled at a particular
time and place, records of an internal robot state or records of swarm-wide
state. To implement a data structure in which robots can retrieve and update
tuples encoding some events, each specific tuple needs a unique identifier
$\tau$ meaningful to all network nodes. In particular, for updates, tuples need
to have a notion of version. For convenience, we achieve versioning by
time-stamping tuples with a global time. Distributed synchronization algorithms
such as vector clocks can be used to implement this aspect.

\section{Methodology}
\label{sec:methodology}

\subsection{Overall Architecture}
\label{ssec:architecture}

\begin{figure}[h!]
    \centering
    \includegraphics[width=1\linewidth]{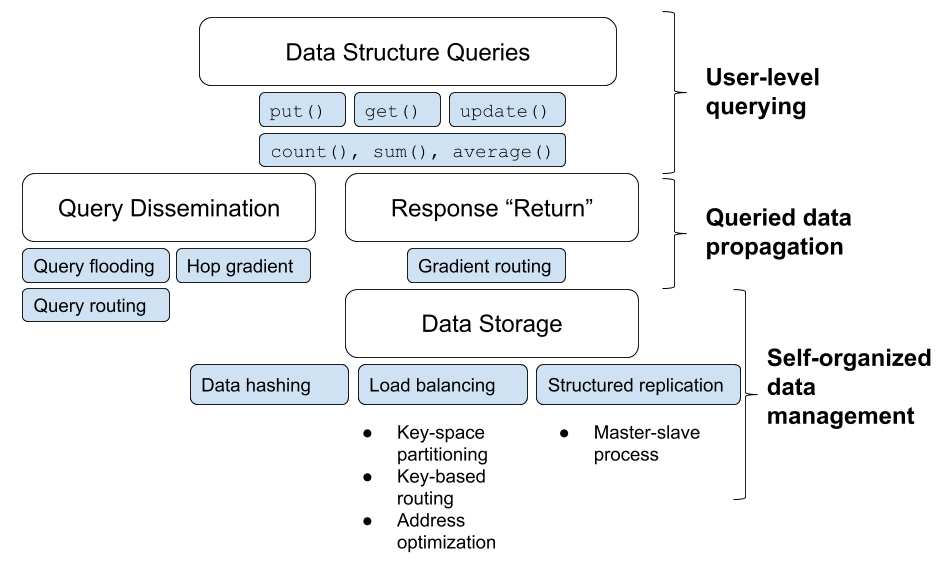}
    \caption{Overall architecture.}
    \label{fig:architecture}
\end{figure}

We describe our design following the structure depicted in Figure
\ref{fig:architecture}. SwarmMesh provides algorithms across three levels of
abstraction.

\subsubsection{User-level querying}
As stated in Section~\ref{sec:introduction}, robots are at the same time the
networking infrastructure and users of the data stored by the network. As a
user, a robot can execute different querying commands on the data
structure. These operations are meant to enable modifying and retrieving
information stored globally as required by the robot behavior. This behavior is
defined by the developer and independent from SwarmMesh.
\subsubsection{Queried data propagation} Another layer of SwarmMesh handles the
dissemination of user read and write queries throughout the data structure. Read
queries are flooded across the network. This type of query requires replies from
certain nodes to be routed back to the robot emitting the query. We route write
operations to a suitable node for storage in a point-to-point fashion.
\subsubsection{Self-organized data management}The bottom layer determines how
the tuples get distributed across nodes. It also ensures a certain degree of
robustness by creating inactive replicas in other nodes in a controlled way. The
main design intuitions driving the data distribution are that: (1) some
events/tuples are more important than others (hierarchy in data hashing); (2)
some nodes are better suited to hold more valuable tuples than others (hierarchy
in key-space partitioning); (3) the hierarchy of nodes changes very often and
should be updated based on local information only.
\subsection{User-level Querying}
\label{ssec:querying}
\begin{figure}
    \centering
    \includegraphics[width=1\linewidth]{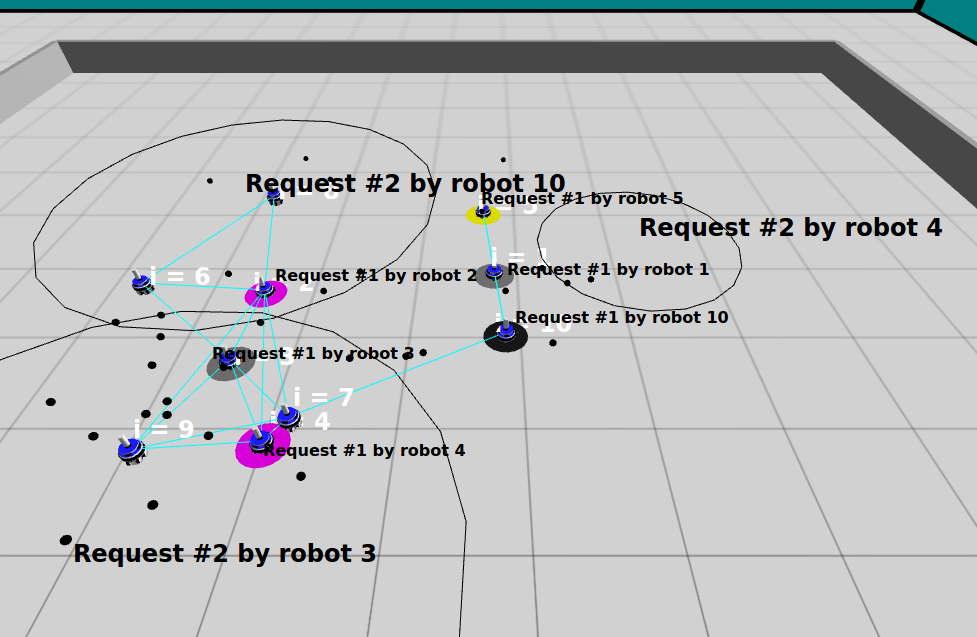}
    \caption{Black dots show locations of events previously written into the
      data structure. Queries of the type \texttt{get(x,y,r)} are drawn with
      black circles representing the area covered. Queries of type
      \texttt{get(k, $\Delta=0$)} are represented by colored disks under the
      query source robot; the color maps to a specific key.}
    \label{fig:queries}
\end{figure}

A robot user can perform the following operations:
\begin{itemize}
\item \texttt{put(k\footnote{The key argument \texttt{k} has an uniquely
      identifying part $\tau$ and a content-dependent part (or hash)
      $\rho_{\tau}$.},v)}: writes a tuple into the data structure. It performs
  an \texttt{erase(k)} to remove any potential outdated version of the tuple and
  a \texttt{store(k,v)} of the new tuple.
\item \texttt{store(k,v)}: assigns a tuple to a particular node in the data
  structure.
\item \texttt{erase(k)}: removes a tuple from the data structure.
\item \texttt{get(k,$\Delta$)}: returns all the values corresponding to keys
  $\in [\,k - \Delta; \, k + \Delta\,]$ (see Figure \ref{fig:queries}).
\item \texttt{get(x,y,r)}: returns all the values for tuples located within a
  radius $r$ of the point $(x,y)$ expressed in a global reference frame. To use
  this feature, we need the added assumption of a global reference frame and the
  ability to locate events in this reference frame (see Figure
  \ref{fig:queries}).
\end{itemize}

Robots can also perform in-network computation:
\begin{itemize}
\item \texttt{count(k,$\Delta$)} or \texttt{count(x,y,r)}: returns the number of
  tuples with keys $\in [\,k - \Delta; \, k + \Delta\,]$ or located within a
  radius $r$ of the point $(x,y)$.
\item \texttt{sum(k,$\Delta$)} and \texttt{sum(x,y,r)}: returns the sum of
  values corresponding to keys $\in [\,k - \Delta; \, k + \Delta\,]$ or located
  within a radius $r$ of the point $(x,y)$.
\item \texttt{average(k,$\Delta$)} or \texttt{average(x,y,r)}: returns the
  result of the corresponding \texttt{count()} and \texttt{sum()} operations as
  a pair.
\item \texttt{min(x,y,r)} or \texttt{min(k,$\Delta$)}: returns the minimum value
  in the associated spatial range or key range.
\item \texttt{max(x,y,r)} or \texttt{max(k,$\Delta$)}: returns the maximum value
  in the associated spatial range or key range.
\end{itemize}

As explained in Section~\ref{ssec:events}, a write operation should be the
result of some local information processing performed by robots in the vicinity
of an event. Existing methods in sensornets can be applied to locally synthesize
low-level sensor readings into a result describing a higher level event such as
source detection \cite{yao1998blind}.
In order to trigger a single data structure write in the vicinity of the event,
we locally elect a leader to perform a \texttt{put(k,v)} operation. The election
logic can be redefined by the developer, although the specifics of this aspect
are beyond the scope of this paper.

The return values of the \texttt{count()}, \texttt{sum()} and \texttt{average()}
operations percolate across nodes. The user robot which emitted the initial
query must combine the intermediate return values into the final
result. Monotonic (i.e., commutative and associative) operations such as
\texttt{min()} and \texttt{max()} do not require combining intermediate results.

The performance of spatial queries, i.e., operations with arguments $(x, y, r)$,
and that of queries by key, i.e., operations with arguments $(k, \Delta)$,
depends heavily on the way we distribute the data in the network. Our approach
is meant to be modular and we present two possible data hashing functions in
Section~\ref{ssec::hashing}.

\subsection{Queried Data Propagation}

\subsubsection{Read-operation flooding} Queries that aim to retrieve data from
the data structure are flooded to all nodes. Each robot emitting a read query
computes the query's unique identifier by concatenating the value of its query
counter and its robot unique identifier.

\subsubsection{Hop gradient}
While flooding the network with a read query, we opportunistically create a
gradient to the source of the query. Upon reception, every robot increments a
hop counter included in the query message and broadcasts it further along. For
each received request, the robot stores the query unique identifier and hop
count in a circular buffer.

\subsubsection{Reply gradient routing}
\label{ssec:gradient}
The hop gradient gives us a convenient way to route replies back to the source
node by forwarding replies from nodes with a higher or equal hop count. For
this, we rely on the assumption that the motion of the robots preserve a
gradient path to the source for long enough, which is a realistic assumption in
most settings when comparing motion speed and information propagation speed.

\subsubsection{Write-operation routing} When a robot writes the result of some
local information processing to the data structure, the tuple may be routed to a
different robot for storage based on its key. This algorithm is described in
Section~\ref{ssec:keyrouting}.

\subsection{Self-organizing Data Management}
\subsubsection{Data Hashing} 
\label{ssec::hashing} When writing a tuple using \texttt{put(k,v)}, the robot
must compute the key \texttt{k}. In our protocol, a key should be in the format
$ k_{\tau} = (\tau, \rho_{\tau})$ where $\tau$ is a tuple unique identifier and
$\rho_{\tau}$ is a value that maps to one or multiple nodes which can store the
tuple.

The robot assigns $\tau$ by concatenating its robot unique identifier and the
count of tuples it has written into the data structure. Each field has a set
number of digits so that every $\tau$ is unique. As stated previously, our
design considers that events vary in importance and we use this property to
distribute them across nodes.

Read queries described in Section~\ref{ssec:querying} can either use
$\texttt{k} = \rho_{\tau}$ or $\texttt{k} = (\tau, \rho_{\tau})$ for tuple
addressing. Queries for $\rho_{\tau}$ can yield multiple tuples while queries in
$(\tau, \rho_{\tau})$ relate to a specific tuple.

The robot computes $\rho_{\tau}$ using a function mapping a characteristic of
the event to its relative importance. We select the function such that the
higher $\rho_{\tau}$, the more valuable the piece of information. We propose two
hashing functions:

\begin{itemize}
\item {Category-based:} We consider that a robot can register different types of
  events. For example, it can mean that the robot has several different on-board
  sensors and determines the event type by the triggered sensor. We use a
  ranking function $R_T(s_{\tau})$ that assigns higher values to event types
  that we consider most important:
  $h_{C} : type_{\tau} \mapsto {\rho}_{\tau} = R_T(type_{\tau})$.
\item{Spatial:} We decide that in a global reference, tuples further away from
  the origin are the most desirable because they are difficult to discover by
  robots. This idea can be generalized to specific areas in any reference frame:
  $ h_{SP} : (x_{\tau},\, y_{\tau}) \mapsto {\rho}_{\tau} = \sqrt{x^2_{\tau} +
    y^2_{\tau}}$.
\end{itemize}


\subsubsection{Key Partitioning}
Similarly to other distributed data structures such as DHTs, nodes partition the
key space to decide which one of them needs to hold tuples corresponding to
specific keys.

As stated in Section~\ref{ssec:architecture}, we use the idea that some nodes
are superior than others.
Our intuition is that a robot with more neighbors $ngbrs_i$ is less likely to
get disconnected from the swarm and is better positioned to dispatch tuples upon
query. A second insight is that the more free data memory a robot has, the less
likely it is to overflow its memory and discard information. We also desire to
have instantaneous self-organized partitioning completely based on local
information.
This leads us to make nodes assign themselves a node identifier $\delta_i$ as follows:
$$ \delta_i(t) =
\begin{cases}
 m_i(t) \cdot ngbrs_i(t) & \text{if } ngbrs_i(t) > 0\\
 1 & \text{otherwise} 
\end{cases}
 $$ 




 A node with node identifier $\delta_i$ can hold a tuple with key
 $(\tau, \rho_{\tau})$ if $\delta_i(t) > \rho_{\tau}$. We refer to this
 condition as (H) in the rest of this text.  In order to store tuples in the
 data structure, we should match the frequency distributions of data hashes and
 node identifiers, i.e., there should be nodes with unique identifiers at least
 high enough to hold the hashed tuples. This has implication on the design of
 the hashing functions. They should map to values smaller than
 $\max_i (M_i) \cdot (N-1)$ and spread the data across likely node identifiers.


\subsubsection{Key-based Routing}
\label{ssec:keyrouting}
If a robot holds one or more tuples not satisfying (H), it places them in a
routing queue. It then tries to send them starting with the highest
$\rho_{\tau}$ to a robot with a high enough node identifier. If there are
candidates satisfying (H) to receive the tuple, the sender picks one at
random. If none of the neighbors satisfy the condition, the robot sends it to
the neighbor of highest $\delta_i$. We impose a limit on the memory capacity
$M_i$ and divide it into routing and storage capacities. In case of overloads on
$M_i$, the robot discards the least important tuple, i.e., with lowest
$\rho_{\tau}$.

\subsubsection{Address Optimization} 
When a robot has an empty routing queue and it stores a tuple with $\rho_{\tau}$
closer to the node identifier of a candidate neighbor, we let the robot evict
the tuple to the corresponding neighbor. This is an optimization to ensure
efficient access to a tuple by key. We further noticed that requiring at least a
half full storage memory helps balancing the load between nodes.



\subsubsection{Structured Replication}
To ensure robustness to node failures, we make copies in neighboring nodes using
a master-slave approach. The master is the robot holding the original tuple. It
picks a slave to hold an inactive copy of the tuple. Robots do not return
inactive tuple copies upon queries; this ensures consistency. Master and slave
exchange a heartbeat signal. If the master fails to receive the heartbeat signal
within a time-out duration, it picks another slave. The master can also send a
kill signal to cancel the inactive copy. The master cancels the copy if the
slave gets outside of a safe radius of communication ($\ll C$) or if it decides
to route the active tuple to another robot. If a slave fails to receive the
heartbeat within the time-out duration from the master, it activates the tuple
copy.


  


\section{Evaluation}
\label{sec:evaluation}

\subsection{Metrics and Parameters}

We evaluated different aspects of our approach such as scalability,
memory-related performance, and routing protocol efficiency.

To study \textit{scalability}, we performed our simulated experiments with 10,
50 and 100 robots uniformly distributed inside an arena sized to impose
densities of 0.6 and 1 robot/$\text{m}^2$. We set the robot communication range
to \unit[2]{m}. These densities imply that the ad-hoc network stays often
connected even with diffuse robot motion. This enables us to study a system
facing intermittent disconnections.

In hash tables, the \emph{load factor} is the number of data items over the
number of memory slots (buckets). This parameter indicates the load of the data
structure and is typically used to decide when to partition of the memory into
an increased number of buckets. For our distributed and self-organized approach,
we define the load factor as $ l_f = \text{number of events} / (N \cdot S)$,
where $S$ is the storage capacity. The memory capacity $M$ includes both storage
and routing capacities.

To understand the performance of the key-based routing algorithm, we track the
number of hops and time steps for a tuple to be routed to a suitable node for
storage and upon query. We implemented messaging with a queue and we imposed a
limit on the bandwidth for outgoing messages.
We only send one tuple at a time.

To study availability, we considered the fraction of tuples received over the
expected tuples for a \texttt{get()} query. We checked for consistency by
confirming that active copies of tuples were all unique.

\begin{table}[t]
  \centering
  \caption{Simulation parameters}
  \label{tab:params}
  \begin{scriptsize}
  \begin{tabular}{|l c|}
    \hline
    Parameter & Value \\
    \hline
    Number of Robots $N$ & $\{10, 50\}$ robots \\
    Communication range $C$ & $2$ m \\
    Memory capacity $M_i$  & 20 tuples $\forall\,\,i$ \\
    Storage capacity $S$ & 10 tuples \\
    Routing capacity $R$ & 10 tuples \\
    Time step & 0.1 s \\
    Bandwidth & 5.7 kB/s \\
    Robot density & $\{0.6, 1\}$ robot/$\text{m}^2$ \\
    Robot speed & $\{0, 5\}$ cm/s \\
    Load factor & $\{0.6, 0.7, 0.8, 0.9, 1\}$ \\
    Event sensing range & 1 m\\
    Event types & 12 \\
    Events generation rate & 5 events/s\\
    Query generation rate & 1 query/s \\
    \hline
  \end{tabular}
  \end{scriptsize}
\end{table}

\subsection{Simulated Experiments}

We tested our system according to the described metrics in the ARGoS multi-robot
simulator \cite{Pinciroli:2012}. We ran simulations with and without robot
motion. We picked a simple diffusion motion with a maximum forward speed of 5
cm/s. For the purpose of testing all available features, robots are equipped
with a range and bearing sensor ($C=$ 2 m), a GPS and a sensor detecting colored
spheres. We disabled line-of-sight obstructions. To materialize events, we put
colored spheres in the environment with each color representing a category of
event. Events were generated in time according to a Poisson distribution and
placed in space according to a uniform distribution.

\subsubsection{Memory-Related Performance}

In our simulations, we allocate limited memory and bandwidth to the data sharing
process. Upon receiving tuples that it can store, a robot progressively fills
its storage memory. The cap on bandwidth combined with the decision to route one
tuple per time step results in some tuples being temporarily placed in a routing
memory. The goal is to keep the storage memory under a value $S$ and the routing
memory under a value $R$. However, we allow either memory to temporarily cross
that threshold provided that the combined memory usage stays under the memory
capacity $M_i$. Any memory overflow leads to robots discarding tuples of lowest
rank. To assess the ability to retain large amounts of information in the data
structure, we generate different numbers of inputs corresponding to load factors
between 0.6 and 1.0. We repeated simulations with and without robot motion and
using either the $h_C$ or $h_{SP}$ hash function. Tables~\ref{tab:mem50}
and~\ref{tab:mem100} show that the fraction of retained tuples is almost always
equal to 1, even with high load factors. This indicates that the collective
memory is properly utilized with robots sharing the data load. In comparison, an
approach that uses full duplication and the same individual memory constraints
would retain $N$ times less tuples (excluding the routing memory).

\begin{table}[t]
  \centering
  \begin{scriptsize}
  \begin{tabular}{llllllll}
    \textbf{load factor} & & & \textbf{.6} & \textbf{.7} & \textbf{.8} & \textbf{.9} & \textbf{1} \\
    \hline
    \hline
    \multirow{6}{*}{\shortstack[l]{category\\ hashing}}
    & \multirow{3}{*}{\shortstack[l]{static\\ topology}}
    &   min  & 1 & 1 & 1 & .996 & .95  \\
    & & mean & 1 & 1 & 1 & .999 & .983 \\
    & & max  & 1 & 1 & 1 & 1     & 1     \\
    \cline{2-8}
    & \multirow{3}{*}{\shortstack[l]{dynamic\\ topology}}
    & min    & 1 & 1 & 1 & .996 & .986 \\
    & & mean & 1 & 1 & 1 & .999 & .992 \\
    & & max  & 1 & 1 & 1 & 1     & 1     \\
    \hline
    \hline
    \multirow{6}{*}{\shortstack[l]{spatial\\ hashing}}
    & \multirow{3}{*}{\shortstack[l]{static\\ topology}}
    & min    & .993 & .932 & .973 & .909 & .758 \\
    & & mean & .999 & .99  & .987 & .948 & .899 \\
    & & max  & 1     & 1     & 1     & .98  & .954 \\
    \cline{2-8}
    & \multirow{3}{*}{\shortstack[l]{dynamic\\ topology}}
    & min    & .997 & .994 & .985 & .984 & .94  \\
    & & mean & .999 & .999 & .996 & .993 & .985 \\
    & & max  & 1     & 1     & 1     & 1     & .998 \\
  \end{tabular}
\end{scriptsize}
  \caption{Tuple retention for $N = 50$ across load factors.}
  \label{tab:mem50}
\end{table}
\begin{table}[t]
  \centering
  \begin{scriptsize}
  \begin{tabular}{lllllll}
    \textbf{load factor} & & & \textbf{.6} & \textbf{.7} & \textbf{.8} & \textbf{.9} \\
    \hline
    \hline
    \multirow{6}{*}{\shortstack[l]{category\\ hashing}}
    & \multirow{3}{*}{\shortstack[l]{static\\ topology}}
    &   min  & 1 & 1 & 1 & .997 \\
    & & mean & 1 & 1 & 1 & .999 \\
    & & max  & 1 & 1 & 1 & 1    \\
    \cline{2-7}
    & \multirow{3}{*}{\shortstack[l]{dynamic\\ topology}}
    &   min  & 1 & 1 & .991 & .977 \\
    & & mean & 1 & 1 & .999 & .996 \\
    & & max  & 1 & 1 & 1    & 1    \\
    \hline
    \hline
    \multirow{6}{*}{\shortstack[l]{spatial\\ hashing}}
    & \multirow{3}{*}{\shortstack[l]{static\\ topology}}
    &   min  & .955 & .912 & .841 & .72  \\
    & & mean & .982 & .97  & .927 & .866 \\
    & & max  & 1    & .999 & .989 & .977 \\
    \cline{2-7}
    & \multirow{3}{*}{\shortstack[l]{dynamic\\ topology}}
    &   min  & .995 & .993 & .994 & .972 \\
    & & mean & .999 & .999 & .998 & .997 \\
    & & max  & 1    & 1    & 1    & 1    \\
  \end{tabular}
\end{scriptsize}
  \caption{Tuple retention for $N = 100$ across load factors.}
  \label{tab:mem100}
\end{table}

\begin{figure*}[h!]
  \centering
    \begin{subfigure}[t]{0.32\textwidth}
      \includegraphics[width=\textwidth]{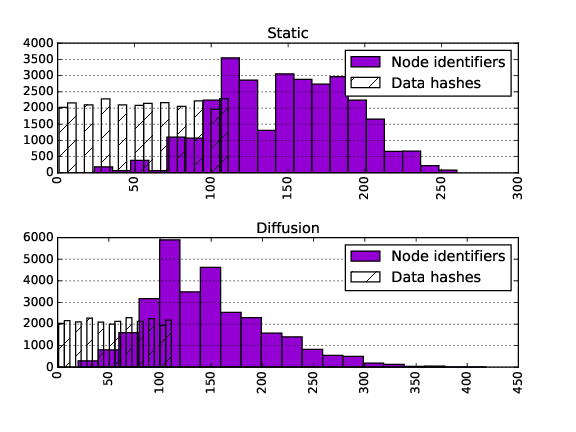}
    \caption{NodeID distribution in category-based hashing.}
    \label{sfig:categoryhist}
  \end{subfigure}
  \begin{subfigure}[t]{0.32\textwidth}
      \includegraphics[width=\textwidth]{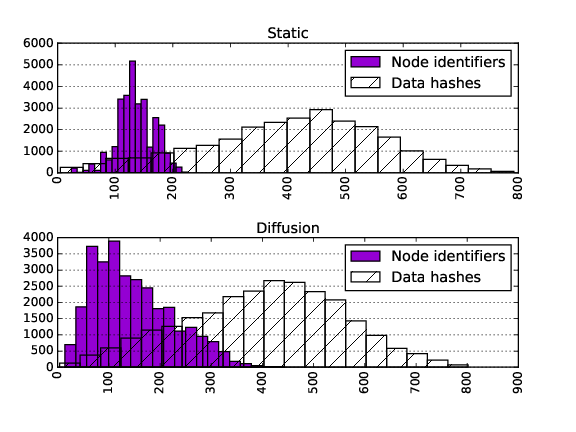}
      \caption{NodeID distribution in spatial hashing.}
      \label{sfig:spatialhist}
  \end{subfigure}
  \begin{subfigure}[t]{0.32\textwidth}
    \includegraphics[width=\textwidth]{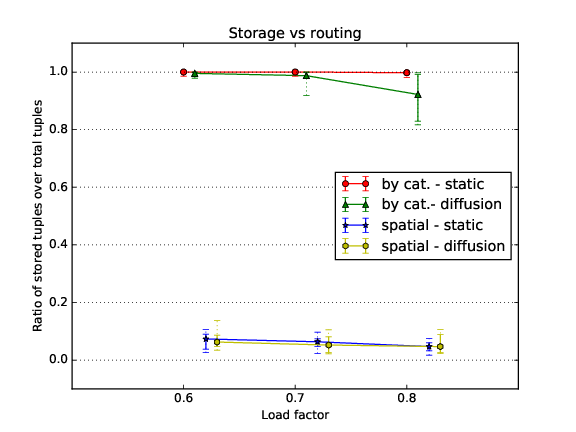}
    \caption{Ratio of stored to routed tuples.}
    \label{sfig:route_store}
  \end{subfigure}
  \caption{Performance of key partitioning with $N=100$ robots.}
\label{fig:partition}
\end{figure*}

In order to evaluate key partitioning, we show histograms of node identifiers
and data hashes across all simulations with 100 robots in Figure
\ref{fig:partition}. In Figure \ref{sfig:categoryhist}, we use the
category-based hash function $h_C$ with a mapping of 12 types of events to
values in $\{ 1, 11, 21, \dots, 121 \}$. We generated the types of events
uniformly in the simulations which naturally leads to the white dashed bar
graphs in Figure \ref{sfig:categoryhist}. Node identifiers are the product of
the current node degree in the communication graph and the node's remaining
storage memory. Therefore, the node identifier distribution depends on a
combination of communication graph topology and load allocation.  Both
situations represented for the category-based have the node identifier
distribution to the right of the data hashes. In the upper graph, robots are
static and their spatial coordinates are sampled in uniform distributions. In
the bottom graph, robots diffuse in an arena sized to impose certain robot
densities (see Table \ref{tab:params}). In Figure \ref{sfig:spatialhist}, we use
the spatial hash function $h_{SP}$ and we show a situation where events are
uniformly generated up to \unit[8]{m} from the origin of the global reference
frame. The function $h_{SP}$ maps the distance in cm to $\rho_{\tau}$ which
yields the Gaussian distribution represented by the white bar graphs. In both
the static and moving case, the node identifier distribution is to the left of
the data keys distribution. This means that, given the key partitioning
condition (H), suitable nodes for storing tuples are scarce or non-existent. As
evidenced by Table~\ref{tab:mem100}, we were still able to retain tuples with
high load factors even in this situation. The reason is that robots shift the
load from their storage memory to their routing memory. This is apparent in
Figure \ref{sfig:route_store} in which the number of tuples in storage memory
normalized by the total number of tuples shows the difference between the use of
$h_{C}$ and $h_{SP}$. With the latter, most tuples remain in routing and bounce
between robots with more free memory. This is not a desirable solution as it
increases the communication overhead. However, it demonstrates a certain
tolerance and seamless adaptation to inappropriate node partitioning. In
practice, with a guess of the environment scale and typical distances, $h_{SP}$
can be scaled so as to provide mapping to a range matching the node identifiers.

\subsubsection{Routing Performance}

\begin{figure*}[h]
  \centering
  \begin{subfigure}[t]{0.32\textwidth}
      \includegraphics[width=\textwidth]{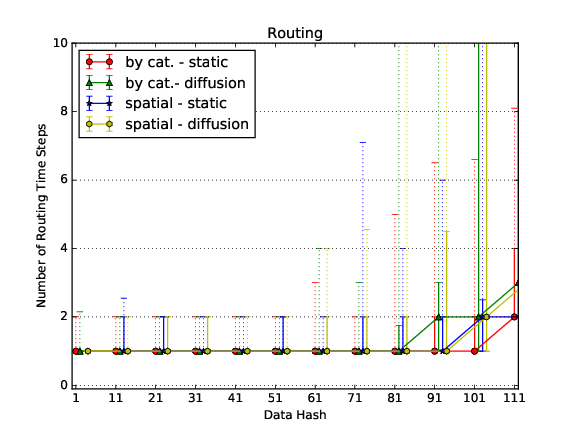}
      \caption{N=10}
  \end{subfigure}
  \begin{subfigure}[t]{0.32\textwidth}
    \includegraphics[width=\textwidth]{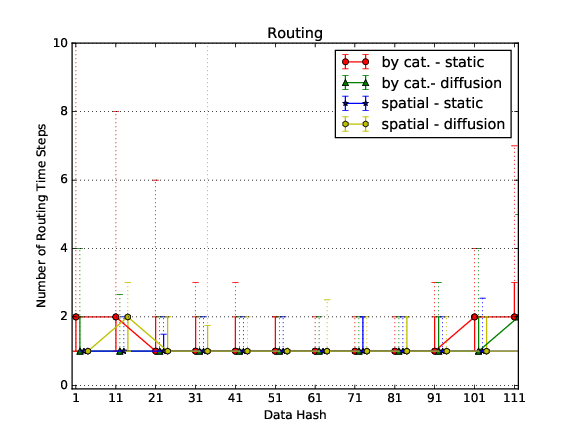}
    \caption{$N=100$}
  \end{subfigure}
  \caption{Number of time steps (\unit[100]{ms} each) for the completion of \texttt{store(k, v)} operations.}
  \label{fig:routing_time}
\end{figure*}

In our approach, routing mechanisms depend on the type of query. A write
operation \texttt{put()} triggers a flooded \texttt{erase()} operation and a
\texttt{store()} propagated through key-based routing (see
Section~\ref{ssec:keyrouting}). The timing for storing a tuple depends on how
difficult it is to reach a suitable node given the tuple
key. Figure~\ref{fig:routing_time} shows the median routing time with 10 and 100
robots for the \texttt{store()} operation. The time tends to increase with the
key indicating that lower range keys find a match faster.

\begin{figure*}[h]
  \centering
  \begin{subfigure}[t]{0.32\textwidth}
      \includegraphics[width=\textwidth]{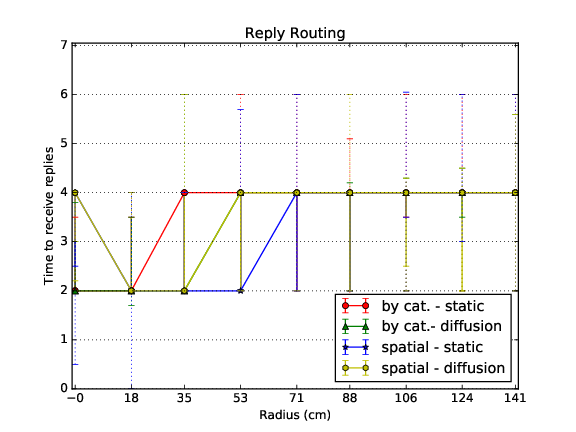}
      \caption{$N=10$}
  \end{subfigure}
  ~
  \begin{subfigure}[t]{0.32\textwidth}
    \includegraphics[width=\textwidth]{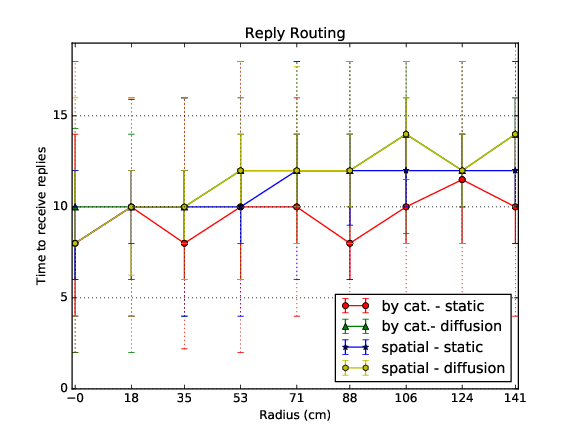}
    \caption{$N=100$}
  \end{subfigure}
  \caption{Number of time steps (\unit[100]{ms} each) for the completion of \texttt{get(x, y, r)} queries.}
  \label{fig:routing_get}
\end{figure*}

Read operations of any type generate a message flooded to all robots. Replies
come back through gradient routing (see Section~\ref{ssec:gradient}). Figure
\ref{fig:routing_get} reports the median duration between a robot emitting a
spatial \texttt{get()} query and receiving the last reply to the query. This
duration tends to increase with the network size and with the radius of the
query.

\subsubsection{Message Load}

\begin{figure*}[h]
  \centering
  \begin{subfigure}[t]{0.32\textwidth}
      \includegraphics[width=\textwidth]{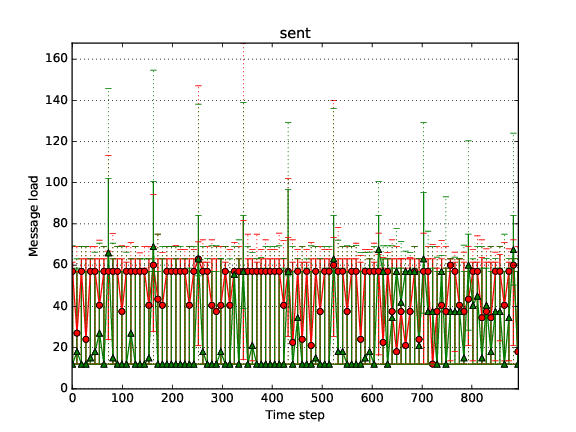}
      \caption{$N=10$}
  \end{subfigure}
  ~
  \begin{subfigure}[t]{0.32\textwidth}
    \includegraphics[width=\textwidth]{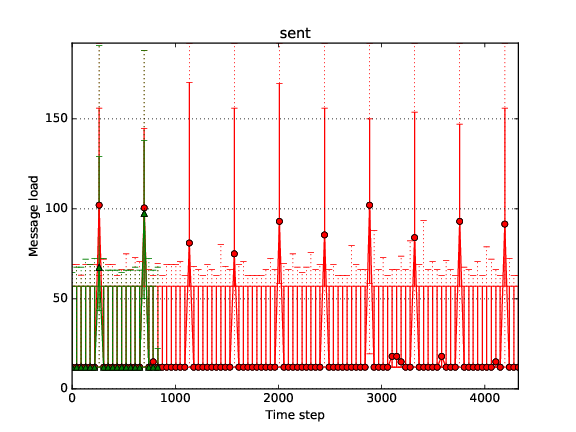}
    \caption{$N=50$}
  \end{subfigure}
  \caption{Bandwidth usage in bytes over time. Time is measured in time steps (\unit[100]{ms}).}
  \label{fig:msgload}
\end{figure*}

The outgoing bandwidth was set to \unit[570]{bytes} per time step for each
robot, with a time step covering \unit[100]{ms}. However, this allowance was
rarely needed. Figure \ref{fig:msgload} shows the median bandwidth usage across
simulations over time, which remains well below the limit we imposed.

\section{Conclusions} 
\label{sec:conclusions}

In this paper we present SwarmMesh, a distributed data structure for low-memory,
low-bandwidth, highly mobile multi-robot systems. The main insight in the design
of SwarmMesh is that the features that characterize the data items play an
important role in deciding where to store the items. Our design is modular, in
that the logic that governs data distribution can be chosen by the user. In this
paper we focus on two methods of distributing storage responsibility, one based
on the category of the data items (for applications in which certain data types
are more important than others), and another based on the position of each data
item. The results of our evaluation show that SwarmMesh displays near-perfect
levels of data retention even for extremely high load factors, adaptively
switching from static storage in the robot memory when load factors are low, to
dynamic storage through frequent data exchange when load factors are severely
high.

Future work involves applying our work to a variety of scenarios, including task
allocation in dynamic environments and collaborative mapping. For the latter
scenario, we will also investigate how to incorporate the size of the data items
as a factor in the data redistribution logic. Finally, our approach lends itself
to privacy and security considerations, whereby the decision on where to store
certain data depends on the reputation of the robots.







\section*{Acknowledgments}
This work was funded by a grant from mRobot Technology Co, Shanghai, China.

\bibliographystyle{IEEEtran}
\bibliography{references}

\end{document}